# Frugal random exploration strategy for shape recognition using statistical geometry

**Authors :** Samuel Hidalgo Caballero[1,2], Alvaro Cassinelli[3], Emmanuel Fort[2], Matthieu Labousse[1]*

**Affiliations :**

[1]Gulliver, CNRS, ESPCI Paris, Université PSL ; 75005 Paris, France, EU.

[2]Institut Langevin, ESPCI Paris, Université PSL ; CNRS, 75005 Paris, France, EU.

[3]School of Creative Media City, City University of Hong Kong; Hong Kong.

*Corresponding author. email: matthieu.Labousse@espci.fr

**Abstract:**

Very distinct strategies can be deployed to recognize and characterize an unknown environment or a shape. A recent and promising approach, especially in robotics, is to reduce the complexity of the exploratory units to a minimum. Here, we show that this frugal strategy can be taken to the extreme by exploiting the power of statistical geometry and introducing new invariant features. We show that an elementary robot devoid of any orientation or observation system, exploring randomly, can access global information about an environment such as the values of the explored area and perimeter. The explored shapes are of arbitrary geometry and may even non-connected. From a dictionary, this most simple robot can thus identify various shapes such as famous monuments and even read a text.

**One-Sentence Summary:**

Blind robots, with no compass and moving randomly, can read and recognize complex shapes based on a statistical approach.

**Main Text:**

Image, object or environment recognition is a highly complex operation which involves acquiring a global information based on spatial correlations. The standard strategy is centralized using multiplexed sensors coupled to complex computational capacities and data processing which can be possibly embarked in high-tech robotic systems. Recently, an alternative fruitful strategy has emerged based on the simple cost-effective basic units collecting unsupervised information, no direct localization capacities and eventually working in parallel in an unmanned manner [1] [2]. This strategy is more robust and well-suited for analysis and exploration with restricted computational and hardware resources, and in the case of harsh or complex environments for which part of the information is not accessible. The most elementary system one can imagine is limited to the acquisition of a local information only with no possible communication, no localization capacity or orientation awareness.

The unavailability of localization capability and the associated randomness of the exploration make the use of statistical geometry crucial to move from local information to a global knowledge of the environment. In this context, the statistical properties of chord length distributions across various kinds of geometrical shapes play a central role for the characterization of size and shape of the intercepted objects with applications such as in acoustics, ecology [3]–[6] image analysis [7], [8], stereology [9]–[11] and reactor design [12]–[15]. The moments of the chord length distributions are related to mathematical invariants and give access to global geometrical parameters such as the volume, the surface or the perimeter of objects. As an historical example, Cauchy formula states that the mean chord length is proportional to the ratio of the volume of an object to its surface area for a 3D object, or to the ratio of its area to its perimeter for a 2D object [16], [17]. Recent generalization to random motion [18]–[23], to closed trajectories [24], [25], and to wave propagation properties [26]–[28] have renewed interest for such invariants. Unfortunately, most of theorems of this field of mathematics are restricted to convex shapes and with no inclusions which restrict the statistical approach for exploration and pattern recognition.

Here, we introduce statistical invariants that extend to any arbitrary, possibly non-convex, non-connected or with inclusions shapes. This generalization allows us to revisit and push the limits of frugal exploration strategies to its most minimalistic version by fully integrating the assets of statistical geometry. We show experimentally that an elementary robot devoid of any localization, orientation and observation systems can assess the surface and perimeter of a region in which it evolves with random displacements. Then, with the help of a dictionary, we show how it recognizes these shapes. We also discuss the benefits of global over sequential exploration strategies in the case of non-connected shapes and apply it to reading strategies.

**Generalized statistical geometry invariants**

Statistical geometry establishes a link between probabilities arising from local properties and global geometric quantities. Cauchy's theorem plays a central role in relating the average chord length of a random distribution $\langle \ell \rangle_{\text{chords}}$ to the ratio of the area $A$ to the perimeter $P$ of the explored domain. The average is realized over all possible chords of the domain. In 2D, it reads $\langle \ell \rangle_{\text{chords}} = \pi A/P$. The Crofton-Hostinský theorem uses the third moment, and in 2D satisfies $\langle \ell^3 \rangle_{\text{chords}} = 3A^2/P$ so that combining these two theorems can yield $P$ and $A$ separately. While Cauchy's theorem can be extended to non-convex shapes [29] unfortunately, the Crofton-Hostinský theorem is only valid for convex shapes, which in practice, prohibits its relevance for exploration and pattern recognition purposes. A generalization of this theorem to arbitrary forms is therefore essential. To this end, we introduce a new statistical invariant. We first label the intersections between a straight trajectory $L$ and the boundaries of the explored shape according to whether they are ingoing $i$ or outgoing $o$ from the shape. We then derive all the distances between pairs of intersections by distinguishing the chords depending on their ends: $l_{io}$, $l_{ii}$ or $l_{oo}$. We define the geometric function of order $n \in \mathbb{N}$ associated to the line $L$, $\mathcal{L}_n(L) = \sum_{i,o} \ell_{io}^n - \sum_o \ell_{oo}^n - \sum_i \ell_{ii}^n$.

We find that the area of the explored shape $A$ satisfies

$$A = \frac{3}{\pi} \frac{\langle \mathcal{L}_3 \rangle_{\text{lines}}}{\langle \mathcal{L}_1 \rangle_{\text{lines}}} \tag{1}$$

with the average being made with respect to all straight lines crossing the shapes (see Supplementary Materials 1 for the demonstration and graphical illustration of Eq. 1). This relation, combined with Cauchy's theorem, allows statistical geometry to be used for exploration and pattern recognition by measuring $P$ and $A$ independently. It applies to any arbitrary shape, whether non-convex, with inclusions, and even non-connected. This establishes an autonomous strategy for obtaining global geometric features without any a priori assumptions.

**Implementation with an elementary robot.**

We experimentally implement this frugal exploration strategy to evaluate the robustness of Eq. 1 under real conditions. We use a commercial toy-robot called *Sphero* which is a self-propelled rolling sphere (see Fig. 1A and Supplementary Materials 2). It possesses an internal clock, two encoded motors for motion and a light sensor. The shape to recognize such as the Eiffel Tower (Fig. 1B), is projected using a video-projector on the arena explored by the robot. The robot performs a series of uncorrelated straight linear motion at 1.0 cm.s$^{-1}$, changing direction randomly as it reaches the edge of the arena (Fig. 1B, see Supplementary Movie 1). Crucially, the robot possesses no information about neither its absolute position nor its direction. From the light received by the sensor, the robot can only detect if it is inside or outside the projected shape. From the times of entries and exits of the shape along the line $L$, the chord lengths are deduced and the functions $\mathcal{L}_1(L)$ and $\mathcal{L}_3(L)$ are calculated on board. From the statistical averages, the area $A$ and the perimeter $P$ of the projected shape are deduced using Eq. 1 and Cauchy's theorem. Note that the memory required to perform the calculations is minimal, containing only the time of intersection events along the latest line to update the previously calculated mean values. Shape recognition is then performed using a dictionary containing a list of area and perimeter entries for a given shape, associated in the present case to famous monuments (Figs 1B-E, see Supplementary Materials 2). Although two different objects can have the same entries, we found in practice that this was hardly ever the case, with up to hundreds of entries. Figure 1B shows the accumulated experimental trajectories during the robot exploring the Eiffel Tower for $N = 10, 100$ and $1000$ random lines. An external camera was used to follow the on-board led which blinks when it is inside of the domain. The final exploration for the other shapes is shown in Figs 1C-E ($N = 1000$). The explorations are found homogenous and statistically isotropic. The chord distribution for the Eiffel Tower is shown in Fig 1F, its average value $\langle \ell \rangle_{\text{chords}}$ and $\frac{3}{\pi}\frac{\langle \mathcal{L}_3 \rangle_{\text{lines}}}{\langle \mathcal{L}_1 \rangle_{\text{lines}}}$ are geometrical invariant (while $\langle \mathcal{L}_1 \rangle_{\text{lines}}$ and $\langle \mathcal{L}_3 \rangle_{\text{lines}}$ are not). The trajectories followed by the estimated parameters and areas as the number of lines $N$ increases (increasing line contrast) can be plotted in a Perimeter-Area representation space for various explored shapes (Fig. 1G, see Supplementary Movie 2). The statistical confidence can be estimated by Monte Carlo simulations for $N = 1000$. The increasingly small ellipses in Fig. 1G show the 75%, 95% and 99% confidence boundaries associated to each shape. This proves that the robots discriminate each registered shape in the dictionary without any statistical ambiguity even though some shapes are very similar, such as in the case of the triangle and the Chichen Itzá pyramid (chosen with an identical area)

The extreme frugality and efficiency of the recognition pathway relies on iterative averaging of the statistical features. Figure 2A shows the distributions of the chord lengths normalized by the maximal length associated to various explored shapes. The full chord length distribution obviously holds some additional geometric information from the shape as shown by distinct profiles. However, in the quest for the simplest exploration strategy, the robot does not need to store all these the data. It only iteratively computes and save the mean values $\langle l \rangle_{\text{chords}}$, $\langle \mathcal{L}_1 \rangle_{\text{lines}}$ and $\langle \mathcal{L}_3 \rangle_{\text{lines}}$. This is performed after each exploration line. Note that even the full knowledge of the chord PDF would not be sufficient to establish a strict correspondence with all possible 2D shapes [30]. For a non-convex shape, a line contains at most $k$ ingoing and outgoing intersection points ($k = 1$ for convex shapes). The robot thus needs to temporarily store at most $k(k-1) + 4$ values. For a complex shape such as the statue of liberty (Fig. 1D), $n \leq 6$ and the robot needs to store at most 34-scalar values. While memory is not restrictive for large robots, it is essential applications at small scales.

We now investigate the convergence to assess the efficiency of the probabilistic method. Figures 2B and 2C show the evolution of the surface area $A$ and the perimeter $P$ respectively as a function of the number of lines $N$. The values are computed by the robot during experiments associated to the five shapes in the shape dictionary (colored contrasted solid lines). The robot is programmed to send these updated geometrical estimates by wireless Bluetooth connection only for visualization purposes as it performs an on-board surface and perimeter estimations. Additionally, we perform Monte Carlo simulations by generating uniformly distributed random lines (colored dimmed solid lines) to compare with the experiments. The convergence towards the expected values of $P$ and $A$ for the experimental and simulated estimates are similar. It is found that only a few hundreds of lines are typically required to measure the area and perimeter with an error of less than 10%. This is sufficient to distinguish the shapes listed in our dictionary. However, in the general case, the recognition speed would strongly depend on the $L_2$-norm in the representation space {Perimeter-Area} between shapes listed in the dictionary. The standard deviation over many independent realizations can be computed from the numerical simulations as a function of exploration lines $N$. Figures 2D and 2E show the evolution of the standard deviations of $A$, $\sigma_A(N)$, and of $P$, $\sigma_P(N)$, respectively for the five shapes of the dictionary. It is found that the convergence, in both cases, is independent of the complexity of the explored shape and scales as $1/\sqrt{N}$. This can be rationalized by noting that the measure converges towards a normal distribution as an ensemble averaging of independent random variables with the same mean and standard deviation. The influence of the shape appears only in the prefactors $\sigma_{0,A}$ and $\sigma_{0,P}$ associated to $A$ and $P$, respectively. The latter increases from the circle, to the square, to the triangle, to the Chichen Itzá pyramid, to the Eiffel Tower, and finally, to the Statue of Liberty. The prefactors thus seem to increase with the complexity of the shape. For regular shapes, complexity can be related to the decrease of the symmetries. For irregular shapes, complexity needs to be assessed quantitatively. The Kullback-Leibler divergence $D_{KL}$ which is a measure of the relative entropy between two probability distributions provides such a parameter. Figure 2F shows the prefactors $\sigma_{0,A}$ and $\sigma_{0,P}$ as a function of the $D_{KL}$ of the chord length distributions associated to the various shapes, with the chord length distributions of a circle taken as a reference. The prefactors increase monotonically with $D_{KL}$ confirming the role of complexity. However, the variation is only a factor of about 4 for a logarithmic variation between distributions of about 3. The dependence on shape complexity is therefore rather modest.

In practice, the exploration must be stopped, once the likelihood of determining a shape is larger than a preset value here 95%. The corresponding likelihood landscape in the representation space {Perimeter-Area} depends on the given dictionary, $\sigma_{0,A}$ and $\sigma_{0,P}$ all of the entries, their relative distance in the {Perimeter-Area} space and the duration of the exploration. Figures 2G and 2I illustrate the influence of the number of crossing lines, N, where each colored domain delineates a 95% likelihood of finding the targeted shape (for N=200 and 1000, see Supplementary Materials 3). Figure 2H and 2J show quantitatively the relative number of lines contributing to a 95% likelihood of recognition. One observes that about a hundred lines are enough to recognize a circular shape with this level of certainty which is due to its isolation from the other shapes in the representation space {Perimeter-Area}. In contrast, it takes more than $10^3$ lines to achieve this level of confidence for the other shapes that are more closely packed in the representation space {Perimeter-Area}.

Eq. 1 remains valid for arbitrary shapes with inclusion or even non-connected shapes. Disconnected objects can thus be explored and gathered within a common global shape with area or perimeter equal to the sum of those of its parts. Since the area and perimeter are geometrical invariants, they are conserved under rigid body transformation such as translations, rotations, mirrors, and any combinations of them (see Supplementary Materials 4). Two alternative strategies can be chosen to recognize unconnected shapes: a global strategy considering all the shapes as a single clustered entity or a local strategy considering the addition of each sub-entity separately. We assess these two strategies through the emblematic case of reading, based on a word or letter-by-letter recognition strategy. The local strategy consists of mapping the capital alphabetic letters into the {Perimeter-Area} representation space which depends on the font. Figure 3A shows the position of the projected capital letters for a Helvetica font and the associated domain where the likelihood of recognition is larger than 95% for N=3 $\times$ $10^4$. Most common fonts do not overlap in the dictionary partly because letters do not transform into each other by symmetry. Figure 3B specifies the number of lines required to distinguish a given letter with a likelihood larger than 95%. Letters can thus be distinguished without ambiguity. For the global strategy, the dictionary is composed of words in the {Perimeter-Area} representation space. Figure 3C shows the dictionary with words sampled from the preamble of the United Nations charter as a dictionary [31] with the associated 95% likelihood domains. All the words can be distinguished. One necessary condition for non-overlapping in the dictionary is the absence of anagrams. In order to compare with the local strategy (Figure 3B), Figure 3D specifies the number of lines, normalized by the of letter in the word, required to distinguish a given letter with a probability larger than 95%. For this moderately-small pool of words, the two strategies are of about the same efficiency. Figures 3E and 3F illustrate the principle of these strategies applied to the word "FREEDOM" with successive identification of each letter (Fig. 3E) and the global estimate (Fig. 3F) of the word perimeter and area. We measure in Fig. 3G and 3H the perimeter and the area of the word FREEDOM. With this moderately-small pool of words, the two strategies are of about the same precision.

The best strategy is a compromise between the size, the choice of the word dictionary and the accumulation of independent errors in the letter-by-letter strategy. On one hand, if the word dictionary is larger than our current one, the words accumulate in a closed region of the Perimeter-Surface space, and as a consequence, a letter-by-letter strategy is more efficient. On the other hand, if we choose a letter-by-letter strategy, the perimeter and surface area computation accumulate the uncertainties corresponding to each letter. In contrast, the direct word recognition combines the chords topology such that the overall noise is smaller. So, for smaller pool of words, measuring directly the word FREEDOM converged quickly by a "word" rather than a letter-by-letter recognition.

The strategy, present here, does not require any knowledge about neither absolute nor relative position as well no information about the direction of motion. On the contrary, the randomness of orientation and absolute position is crucial. At every single stage of the physical computing, the robots acquire no global information about the number of holes, the convexity or any global information. However, by being ignorant about any global information and by measuring only local quantities, the robot is able to recognize shapes listed in a dictionary by measuring the perimeter and the area of any given 2D shapes. This was only possible thanks to the introduction of a new generalized statistical invariant which allows us to estimate the area from a random exploration of any 2D domain, regardless of its complexity. Finally, we note that the dictionary is not a necessary prerequisite and is used here for convenience. A pre-classification phase, for example with k-means or Gaussian mixture strategies, can be added to our protocol to perform an autonomous and unsupervised recognition. Extending our technique to a swarm of robots, to take advantage of distributive computation strategies, is a natural step where each can compute these invariants locally and share them with the others to improve convergence. However, the inevitable contact between individuals and the possible formation of aggregates makes swarm computation a real challenge, for which new invariants for more disordered paths need to be discovered.

**Funding:**

European Union's Horizon 2020 (Marie Skłodowska-Curie grant No 754387)

AXA Research Fund

CONACYT Mexico (mobility Grant CVU-849369)

French National Research Agency (ANR-10-LABX-24).

**Author contributions:**

Conceptualization: EF, ML

Methodology: SHC, AC, EF, ML

Investigation: SHC, AC, EF, ML

Visualization: ML, EF, SHC

Funding acquisition: ML

Project administration: ML, EF

Supervision: ML, EF

Writing – original draft: ML, EF

Writing – review & editing: ML, EF, SHC, AC

**Competing interests:** Authors declare that they have no competing interests.

**Data and materials availability:** All data are available in the main text or the supplementary materials.

**Supplementary Materials**

Supplementary Text 1-5

Figs. S1 to S2

Tables S1

References ([15])

Movies SM1 to SM2

**Figure 1**

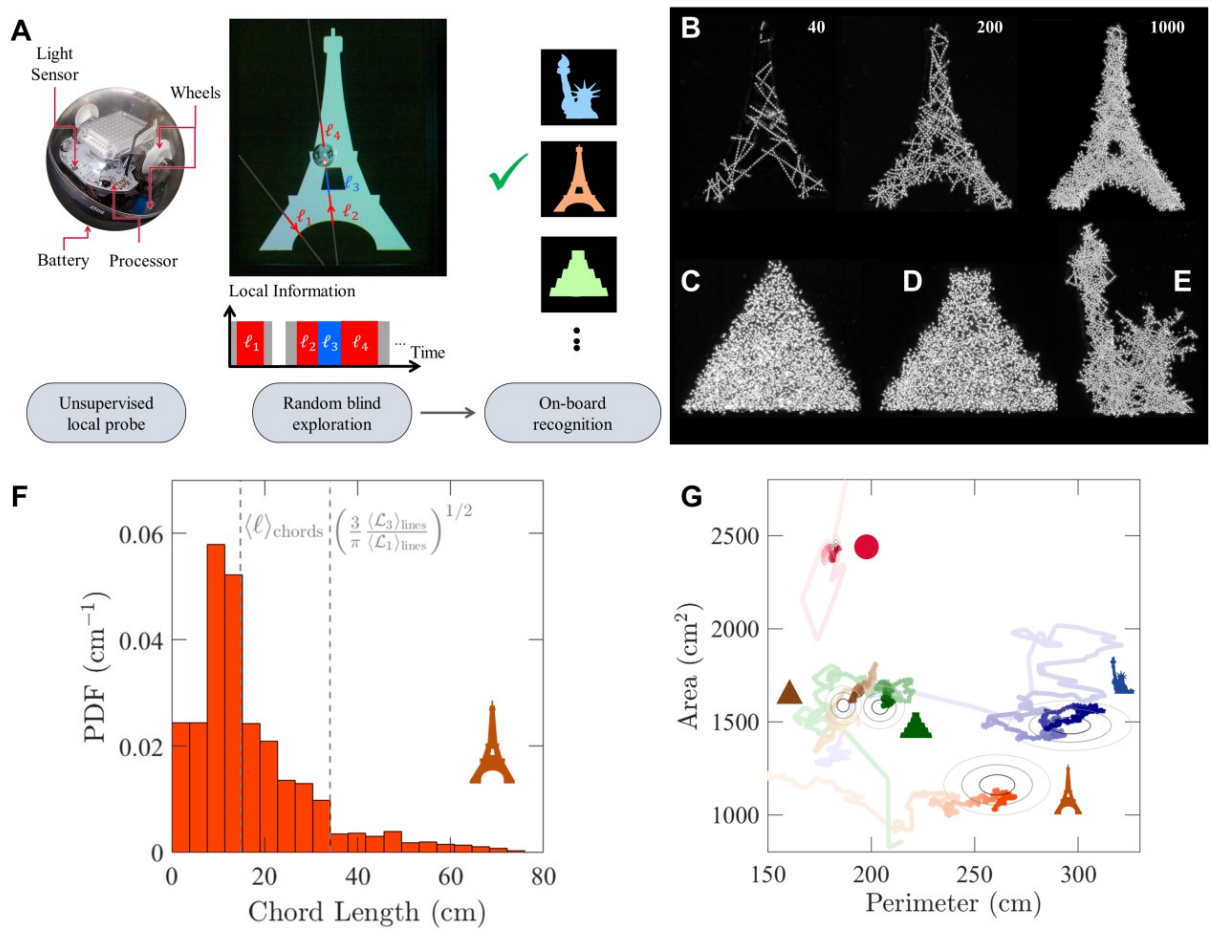

**Fig. 1. Principle and implementation of robotic shape recognition by random exploration.** (A) Photo and structure of the autonomous *Sphero* toy-robot. (B)-(E) Exploration of a projected Eiffel Tower (B) shape by random straight-line motion acquiring only the chord lengths. By using statistical geometry to compute invariant quantities and shape parameters such as the perimeter $P$ and the area $A$ of the explored shape enables recognition from a dictionary. Snapshots of accumulated experimental light blinks emitted by the robot during its exploration of the Eiffel Tower for $N = 40, 200$ and $1000$ lines (B) and for a triangle (C), the Chichen Itzá pyramid (D), and the Statue of Liberty for $N = 1000$ lines (E). (F) Chord length distribution for the Eiffel Tower shape with the mean chord length invariant $\langle \ell \rangle_{chords}$ and $A^{1/2} = \left(\frac{3}{\pi} \frac{<\mathcal{L}_3>_{lines}}{<\mathcal{L}_1>_{lines}}\right)^{1/2}$ for comparison. An extra tilde indicates that the length-scale is normalized by the maximum measured chord of a given shape. (G) Trajectories followed by the estimated parameters $(P, A)$ as the number of lines $N$ increases (increasing line contrast) in the Perimeter-Area representation space for various explored shapes. 75%, 95% and 99% statistical confidence limits associated to each shape estimated by Monte Carlo simulations for $N = 1000$ (ellipses with increasing contrast and decreasing size).

**Figure 2**

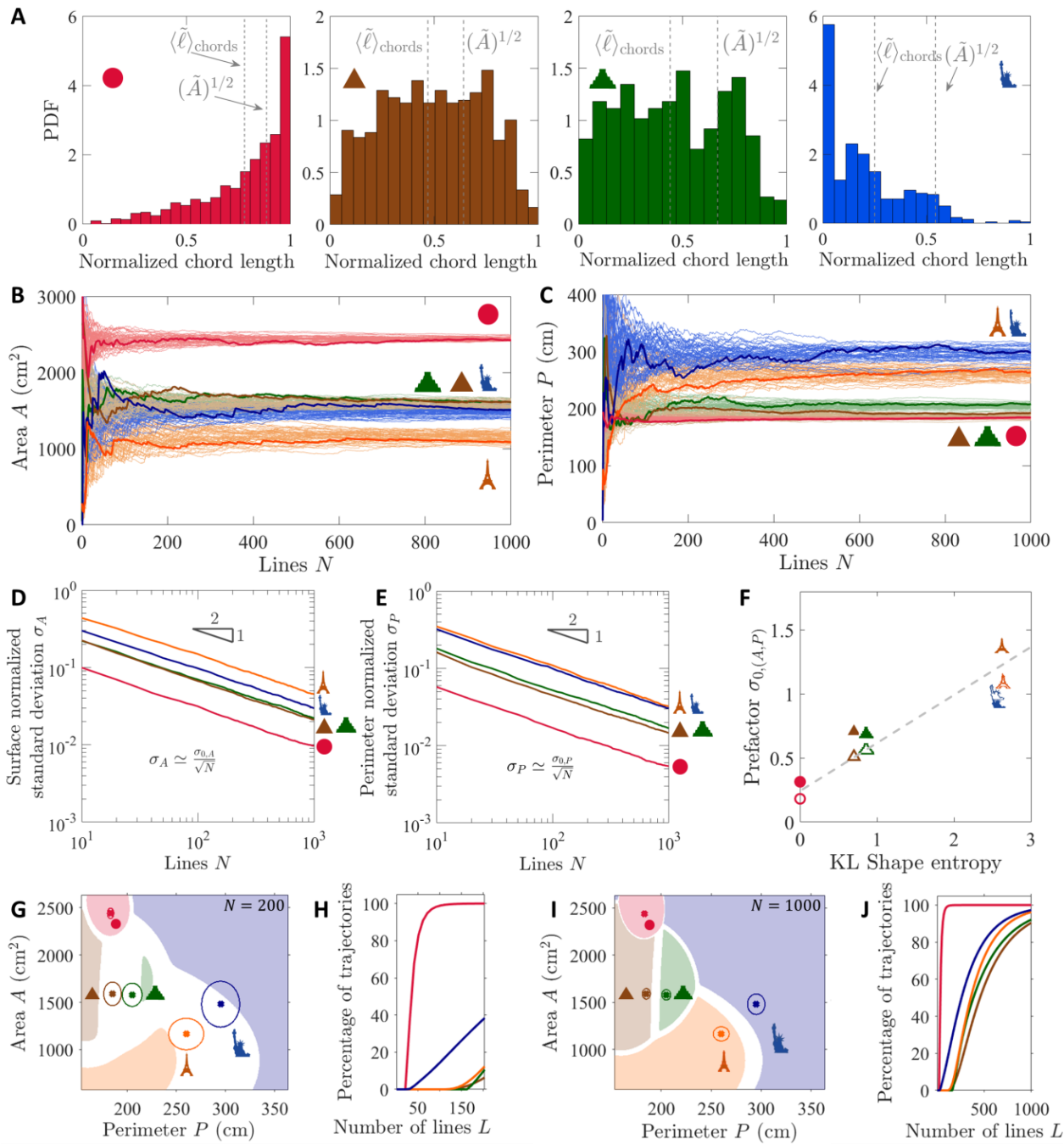

**Fig. 2. Convergence properties.** Common color code: Circular shape (pink), Chichén Itzá pyramid (green), triangle (brown), Statue of Liberty (blue), Eiffel tower (orange) **(A)** Histogram of chords for the four remaining shapes. The normalized mean chord length invariant $\langle \ell \rangle_{\text{chords}}$ and $\tilde{A}^{1/2} = \left(\frac{3}{\pi}\frac{<\tilde{\mathcal{L}}_3>_{\text{lines}}}{<\tilde{\mathcal{L}}_1>_{\text{lines}}}\right)^{1/2}$ for comparison. Lengths are normalized by the maximum chord length for each shape and is indicated by a tilde sign. **(B) & (C)** Evolution of the area and perimeter estimated autonomously by the robot during its exploration. Thick lines correspond to an experimental realization, thinner lines correspond to Monte Carlo simulations and give an indication of the reduction of the fluctuations over different realizations. **(D) & (E)** Evolution of the standard deviation of the surface and perimeter computed from 1000 independent Monte Carlo simulation as the function the number of lines and follows a ½ power law convergence $\sigma_A \simeq \sigma_{A,0}/\sqrt{N}$ and $\sigma_P \simeq \sigma_{P,0}/\sqrt{N}$. **(F)** Evolution of the prefactor $\sigma_{A,0}$ or $\sigma_{P,0}$ with the shape entropy of the domain defined as the Kullback-Leibler divergence (KL) of the chord distribution. The dashed grey line is a guide for the eye. The chord distribution of a circular shape is taken as a reference for the KL calculation.

Figure 3

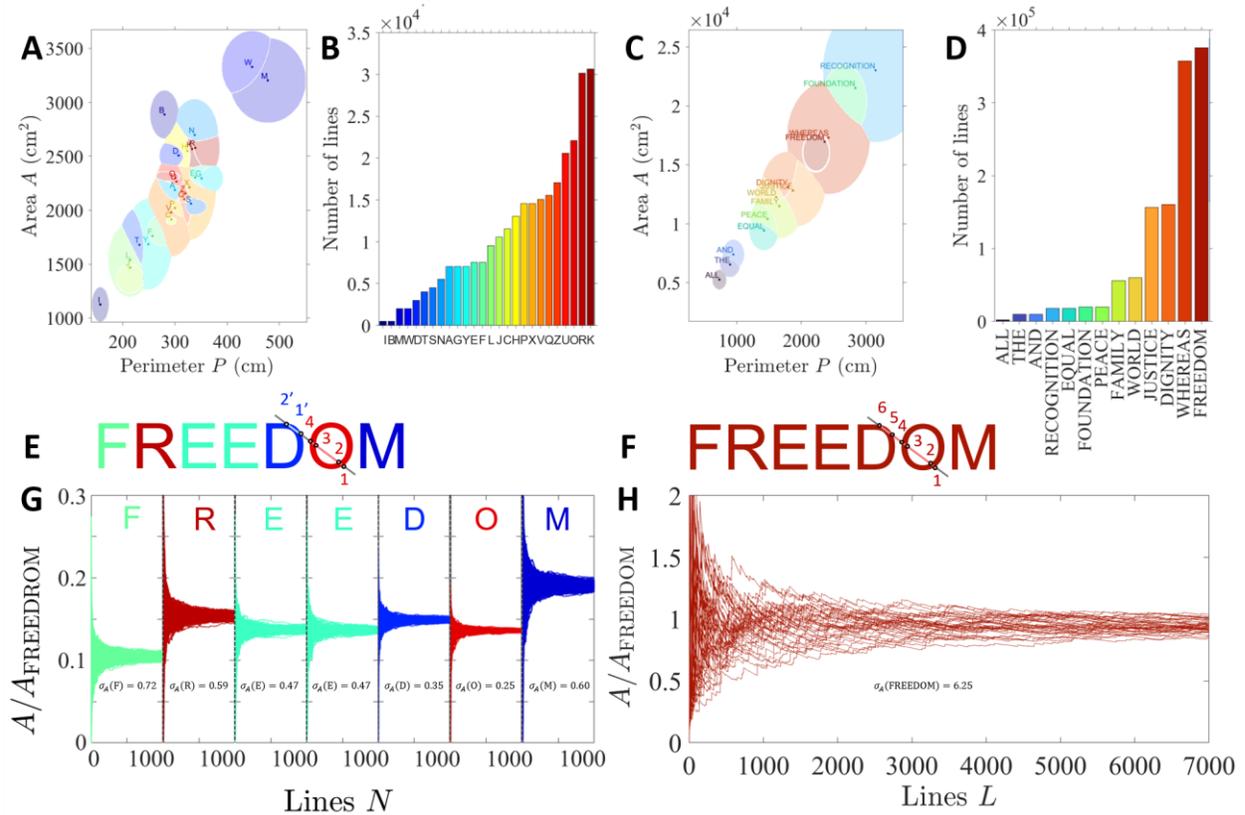

**Fig. 3 Global versus local exploration strategies applied to reading.** **(A)** Likelihood landscape in the {perimeter-area} representation for the projected capital letters of the alphabet with a likelihood larger than 95%. **(B)** Specifies the number of lines required to distinguish a given letter without ambiguity, in this case with more than 95% of probability. **(C-D)** Global strategy. The dictionary is composed of words in the {Perimeter-Area} representation space. **(C)** Words sampled from the preamble of the United Nations charter as a dictionary, [31] with the associated 95% likelihood domains. The words can be distinguished if there is an absence of anagrams. **(D)** number of lines, normalized by the of letter in the word, required to distinguish a given letter with a probability larger than 95%. **(E) – (F)** Comparison of both strategies applied to the word "FREEDOM" with successive identification of each letter **(E)** and the global estimate **(F)** of the word perimeter and area. **(G) - (H)** Area estimation of the word FREEDOM and its associated error, with a local **(G)** or a global **(H)** strategy. With this moderately-small pool of words, the two strategies are of about the same precision.